\documentclass[10pt,twocolumn,letterpaper]{article}

\usepackage[pagenumbers]{cvpr}   

\usepackage{listings}

\lstdefinestyle{promptstyle}{
  basicstyle=\ttfamily\scriptsize,
  breaklines=true,
  breakatwhitespace=false,
  columns=fullflexible,
  keepspaces=true,
  showstringspaces=false,
  frame=none,
  xleftmargin=1em,
  xrightmargin=1em,
  aboveskip=0.6em,
  belowskip=0.6em
}

\newcommand{\promptheading}[1]{%
  \vspace{0.8em}
  \noindent\textbf{#1}\par
  \vspace{0.2em}
}

\usepackage{multirow}

\definecolor{cvprblue}{rgb}{0.21,0.49,0.74}
\usepackage[pagebackref,breaklinks,colorlinks,allcolors=cvprblue]{hyperref}

\def\paperID{0}
\def\confName{CVPR}
\def\confYear{2026}

\title{Answer Self-Consistency with Margin-Triggered Question Re-Arbitration for the CVPR 2026 VidLLMs Challenge}

\author{
Tomoya Miyazawa \quad Hiroyasu Okuno\\
Data Analytics Labo Co.\\
{\tt\small \{tomoya.miyazawa, hiroyasu.okuno\}@dalab.jp}
}

\begin{document}
\maketitle

\begin{abstract}
In this report, we present our solution for Track 2 of the CVPR 2026 VidLLMs Challenge. 
This track evaluates visual relational reasoning in videos, where models must infer relations that are not always explicitly visible. 
We propose \emph{Answer Self-Consistency with Margin-Triggered Question Re-Arbitration} (ASC-MQRA), a training-free test-time reasoning framework built on a multimodal reasoning model. 
The core ASC component performs multiple stochastic video question-answering runs and aggregates their answer choices through answer-level self-consistency. 
This substantially improves over single-pass inference and forms our final test submission.

We further study MQRA, a conditional re-arbitration module for low-margin examples where the first-stage vote distribution indicates uncertainty. 
Our vote-margin analysis shows that low-margin examples often retain the ground-truth answer among the top candidates, motivating MQRA to narrow the candidate set and re-watch the video only over the retained candidates. 
On validation, MQRA further improves over ASC, indicating that low-margin vote distributions can provide a useful uncertainty signal.
On test, however, MQRA slightly degrades performance relative to ASC, suggesting that re-arbitration is sensitive to the size and category distribution of the triggered subset.
Our final test submission therefore uses ASC without re-arbitration, achieving 72.73 average accuracy and 78.34 category-wise macro average accuracy on validation, and 81.16 average accuracy and 80.91 category-wise macro average accuracy on test. 
This report details our prompting strategy, implementation setup, ablation studies, and diagnostic analyses. 
The code is available at \url{https://github.com/data-analytics-labo/ASC-MQRA}.
\end{abstract}
\section{Introduction}
\label{sec:introduction}

Video question answering has recently benefited from increasingly capable Video Large Language Models (Video LLMs). 
However, Track 2 of the CVPR 2026 VidLLMs Challenge requires a different form of understanding: the answer is often not directly visible in a single frame or shot~\cite{vidllms2026challenge}. 
Instead, models must integrate implicit cues distributed across the video, such as object continuity, motion direction, spatial layout, temporal order, and contextual implications. 
This makes the task less about isolated visual recognition and more about reasoning over incomplete and discontinuous evidence.

Track 2 is based on Visual Relational Reasoning Question Answering (VRR-QA), a benchmark designed to evaluate visual relational reasoning beyond explicitly visible cues~\cite{swetha2025vrrqa,implicitqa_dataset}. 
VRR-QA highlights a key limitation of current video question-answering systems: they often struggle when the decisive relation is only implied across frames rather than directly shown. 
For example, solving a question may require tracking the same entity across shot changes, inferring relative motion from indirect cues, or reconstructing spatial relations from separated observations. 
These properties make VRR-QA a natural setting for inference-time reasoning strategies.

In this work, we develop a training-free solution built on Gemini 3.1 Pro Preview. 
Because the challenge does not restrict the model family, we use an off-the-shelf multimodal model and focus on improving the inference pipeline rather than training a task-specific video-language model. 
Preliminary experiments showed that predictions were not fully stable across repeated runs, even under low-temperature decoding settings. 
This motivated us to use test-time scaling: sampling multiple reasoning attempts and aggregating their final answer choices.

Our core method is answer-level self-consistency, inspired by self-consistency decoding for reasoning~\cite{wang2023selfconsistency}. 
For each question, we run the video question-answering prompt multiple times with stochastic decoding and select the final answer by majority vote. 
Although each run produces video-grounded evidence, the aggregation itself uses only the predicted answer choice, avoiding reliance on noisy intermediate explanations.

We further investigate \emph{Margin-Triggered Question Re-Arbitration} (MQRA), a conditional refinement for low-margin examples. 
When the first-stage vote distribution is split, we retain the top candidate answers and ask the model to re-watch the video with a prompt focused on distinguishing the remaining candidates. 
This allocates additional inference only to contested examples where the correct answer may still be recoverable, although its effectiveness depends on the triggered examples.

Our main contributions are summarized as follows:
\begin{itemize}
    \item We present a training-free solution for Track 2 of the CVPR 2026 VidLLMs Challenge based on test-time scaling of a multimodal reasoning model.
    \item We show that stochastic answer-level self-consistency improves robustness over single-pass video question answering.
    \item We study margin-triggered question re-arbitration as a conditional refinement for low-margin examples and analyze when it helps or hurts.
    \item We provide ablation studies on inference configuration, answer-level self-consistency, re-arbitration thresholds, and cross-model behavior, together with diagnostic analyses of recoverable uncertainty.
\end{itemize}
\section{Benchmark Overview}
\label{sec:benchmark}

Track 2 of the CVPR 2026 VidLLMs Challenge is based on \emph{Visual Relational Reasoning in Videos Beyond Explicit Cues} (VRR-QA), a multiple-choice video question answering benchmark for evaluating implicit visual relational reasoning~\cite{swetha2025vrrqa,implicitqa_dataset}. 
Unlike standard video question answering benchmarks that often focus on objects, actions, and events directly observable in individual frames or short clips, VRR-QA targets hidden relationships and contextual cues that are only implied across frames. 
Such questions may require inferring entity continuity, motion direction, spatial layout, temporal order, physical context, or social relationships from incomplete visual evidence.

VRR-QA contains 1K expert-annotated question-answer pairs from 1K cinematic video clips sourced from movies. 
The benchmark covers diverse genres, time periods, and visual styles, including both live-action and animated content. 
This cinematic setting is challenging because important relations are often conveyed through off-screen implications, omitted events, viewpoint shifts, or indirect cues rather than explicit visual co-occurrence.

The dataset is organized into nine implicit reasoning categories: causal and motivational reasoning, inferred counting, lateral spatial reasoning, motion and trajectory dynamics, physical and environmental context, relative depth and proximity, social interaction and relationships, vertical spatial reasoning, and viewpoint and visibility. 
Each example consists of a video clip, a question, and multiple answer choices, and the system must output a single answer choice. 
The leaderboard reports average accuracy and category-wise macro average accuracy. 
In our analysis, official category labels are available for the validation split, while test category labels are not released to participants.
\section{Method}
\label{sec:method}

\subsection{Overview}

We propose a training-free test-time inference pipeline for multiple-choice video question answering. 
Given a video clip, a question, and a set of answer choices, our method selects the single best answer without task-specific training or fine-tuning. 
The pipeline consists of two components: \emph{Answer Self-Consistency} (ASC), which aggregates multiple independent watch-and-answer runs, and \emph{Margin-Triggered Question Re-Arbitration} (MQRA), which allocates additional inference to low-margin examples.

\begin{figure}[t]
    \centering
    \includegraphics[width=0.95\linewidth]{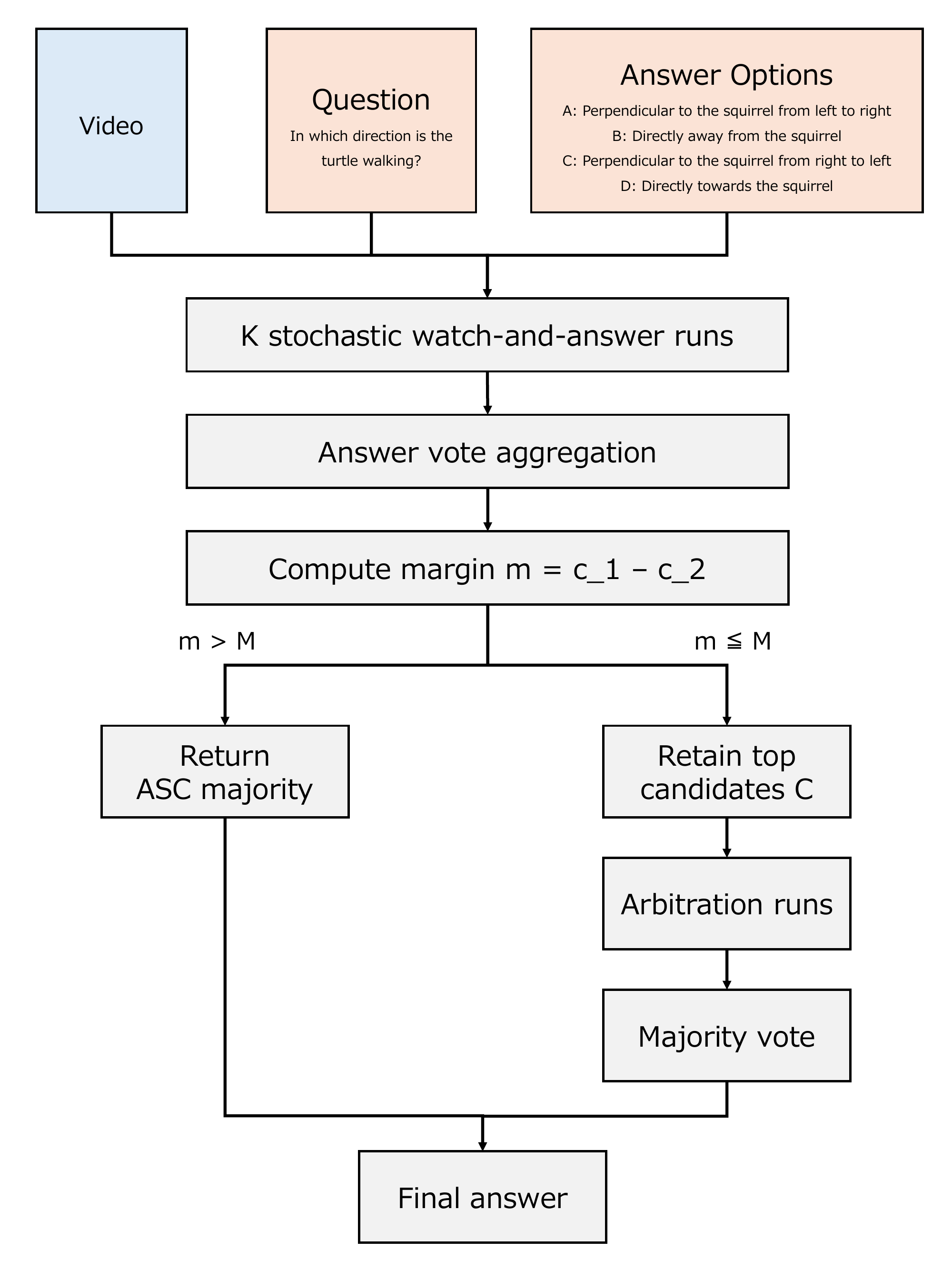}
    \caption{
    Overview of the ASC-MQRA pipeline. 
    Multiple stochastic watch-and-answer runs are first aggregated by answer-level self-consistency. 
    Low-margin examples are then re-arbitrated over retained candidate answers.
    }
    \label{fig:method_overview}
\end{figure}

Figure~\ref{fig:method_overview} illustrates the overall pipeline. 
We first run the same video question-answering prompt multiple times over all answer choices. 
Each run outputs an answer choice and supporting video-grounded evidence, but only the answer choices are used for aggregation. 
If the first-stage vote margin is sufficiently large, the majority answer is returned. 
Otherwise, the system retains the top candidate answers and performs focused re-arbitration by re-watching the video with only those candidates.

\subsection{Answer Self-Consistency}

Our core component is answer-level self-consistency, inspired by self-consistency decoding for reasoning~\cite{wang2023selfconsistency}. 
For each example, we run the same watch-and-answer prompt $K$ times with stochastic decoding. 
Each run receives the video $v$, question $q$, and all answer choices $\mathcal{A}$, and returns an answer choice $y_i \in \mathcal{A}$ and supporting evidence $e_i$:
\[
    y_i, e_i = f_{\theta}(v, q, \mathcal{A}), \quad i = 1,\ldots,K.
\]

The prompt asks the model to answer only from visible or strongly implied video evidence, while noting that the decisive cue may involve distant events, subtle state changes, or entity continuity across shots. 
The evidence field is included to encourage grounded reasoning within each individual run, but it is not used for aggregation. 
ASC aggregates only the predicted answer choices:
\[
    \hat{y}_{\mathrm{ASC}} =
    \arg\max_{a \in \mathcal{A}}
    \sum_{i=1}^{K} \mathbf{1}[y_i = a].
\]
If multiple choices receive the same maximum number of votes, we randomly break the tie among the tied choices. 
This answer-level aggregation avoids relying on noisy intermediate explanations, which we found to be unstable for difficult implicit reasoning questions.

\subsection{Margin-Triggered Question Re-Arbitration}

ASC improves robustness, but a small vote margin indicates that the prediction remains contested. 
We therefore study MQRA, a conditional refinement module that reconsiders only low-margin examples. 
Let $c_1$ and $c_2$ be the largest and second-largest vote counts after ASC. 
We define the vote margin as
\[
    m = c_1 - c_2.
\]
If $m > M$, where $M$ is a predefined threshold, we return the ASC majority answer. 
If $m \leq M$, the example is sent to re-arbitration.

For re-arbitration, we retain the top candidates from the first-stage vote distribution. 
In the standard case, this corresponds to the two most-voted choices; when multiple choices are tied at the second-highest count, all tied choices are retained. 
Let $\mathcal{C} \subseteq \mathcal{A}$ denote the retained candidate set, and let $C = |\mathcal{C}|$ be the number of retained candidates.

The arbitration prompt receives only the original video, the question, and the retained candidates. 
It is not given the first-stage vote counts or first-stage generated evidence, which avoids amplifying noisy intermediate explanations. 
Instead, the prompt asks the model to clarify the question, identify the decisive difference between the retained candidates, and re-watch the video before choosing one candidate. 
For each low-margin example, we run the arbitration prompt
\[
    R = 2C + 1
\]
times and aggregate the arbitration answers by majority vote:
\[
    \hat{y}_{\mathrm{MQRA}} =
    \arg\max_{a \in \mathcal{C}}
    \sum_{j=1}^{R} \mathbf{1}[z_j = a],
\]
where $z_j$ is the answer choice from the $j$-th arbitration run. 
The final prediction is
\[
\hat{y} =
\begin{cases}
\hat{y}_{\mathrm{ASC}}, & \text{if } m > M, \\
\hat{y}_{\mathrm{MQRA}}, & \text{if } m \leq M.
\end{cases}
\]

\subsection{Implementation Details}

We implemented the inference workflow using LangGraph~\cite{langgraph}, which allowed us to express the ASC stage, margin check, candidate retention, and optional MQRA stage as a stateful graph. 
We used W\&B Weave~\cite{weave} to trace model inputs, outputs, intermediate predictions, vote distributions, and final decisions across experiments. 
These tools were used only for workflow orchestration and experiment tracking; they did not change the prompting strategy, model outputs, or aggregation rules described above.
\section{Experiments}
\label{sec:experiments}

We evaluate our method on the official validation and test splits of Track 2. 
We report two metrics: \emph{AvgAcc}, the instance-level average accuracy over all questions, and \emph{MacroAvgAcc}, the average of category-wise accuracies across reasoning categories. 
All scores are reported as percentages. 
Unless otherwise specified, each prediction is a single answer choice in the official multiple-choice format.

\subsection{Experimental Setup}

We use Gemini 3.1 Pro Preview as the backbone multimodal model. 
For each example, the model receives the video clip, the question, and the answer choices, and returns a structured response from which we extract the predicted answer choice. 
We tune basic inference configurations on the validation split and use the selected configuration for subsequent experiments.

Our default setting uses the medium thinking level and a video sampling rate of $4$ fps. 
For Answer Self-Consistency (ASC), we use $K=10$ watch-and-answer runs with temperature $1.0$ unless otherwise specified. 
For Margin-Triggered Question Re-Arbitration (MQRA), we vary the margin threshold $M$ and use $R=2C+1$ arbitration runs, where $C$ is the number of retained candidate answers.

Official category labels are available for the validation split, while test category labels are not released to participants. 
Therefore, category-wise analysis is mainly conducted on validation, and test results are reported using the official leaderboard metrics.

\subsection{Main Results}

Table~\ref{tab:main_results} summarizes the main results. 
Single-pass inference provides a strong baseline, but ASC improves performance by aggregating multiple watch-and-answer runs. 
ASC-MQRA further improves validation performance for low-margin settings: $M=0$ and $M=1$ tie on validation AvgAcc, and $M=0$ achieves the best validation MacroAvgAcc. 
However, increasing the margin threshold does not always improve performance, and the effect on the smaller test split is more sensitive to the number and distribution of triggered examples.

Although MQRA improves validation performance, our final leaderboard submission uses ASC without re-arbitration, which achieves the best test performance among our evaluated variants.

\begin{table}[t]
\centering
\caption{
Main results on the official validation and test splits. 
Single-pass scores are averaged over 5 runs.
ASC denotes answer self-consistency over multiple watch-and-answer runs. 
MQRA denotes margin-triggered question re-arbitration for examples satisfying $c_1 - c_2 \leq M$.
Our final leaderboard submission corresponds to ASC, $K=10$.
}
\label{tab:main_results}
\resizebox{\linewidth}{!}{
\begin{tabular}{lcccc}
\toprule
Method
& Val AvgAcc
& Val MacroAvgAcc
& Test AvgAcc
& Test MacroAvgAcc \\
\midrule
Single pass
& 68.41
& 74.60
& 72.71
& 73.62 \\
ASC, $K=10$ (final)
& 72.73
& 78.34
& \textbf{81.16}
& \textbf{80.91} \\
ASC-MQRA, $M=0$
& \textbf{73.23}
& \textbf{78.69}
& 80.85
& 80.33 \\
ASC-MQRA, $M=1$
& \textbf{73.23}
& 78.48
& 80.85
& 80.33 \\
ASC-MQRA, $M=2$
& 72.83
& 78.28
& 78.09
& 77.36 \\
\bottomrule
\end{tabular}
}
\end{table}

\subsection{Ablation Studies}

We conduct additional ablation studies on the basic inference configuration and the decoding temperature used for ASC. 
We select the default configuration based primarily on validation performance, since the test split is smaller and more sensitive to run-to-run variation.

\paragraph{Inference configuration.}
Table~\ref{tab:inference_config} shows the effect of thinking level and video sampling rate. 
The thinking level controls the amount of reasoning allocated by the model, while the sampling rate controls the temporal density of the video input. 
Among the tested settings, the medium thinking level and $4$ fps provide the best validation performance, and we use this configuration as the default.

\begin{table}[t]
\centering
\caption{
Ablation of basic inference configurations on the validation split. 
Scores are reported as percentages and shown as the mean $\pm$ standard deviation over 5 runs.
}
\label{tab:inference_config}
\resizebox{\linewidth}{!}{
\begin{tabular}{llcc}
\toprule
Factor
& Setting
& Val AvgAcc
& Val MacroAvgAcc \\
\midrule
Thinking level
& Low
& 67.87 $\pm$ 0.38
& 73.53 $\pm$ 0.52 \\
Thinking level
& Medium
& \textbf{68.41} $\pm$ 0.85
& \textbf{74.60} $\pm$ 0.99 \\
Thinking level
& High
& 67.29 $\pm$ 0.39
& 73.50 $\pm$ 0.38 \\
\midrule
Sampling rate
& 1 fps
& 65.53 $\pm$ 0.18
& 71.13 $\pm$ 0.54 \\
Sampling rate
& 4 fps
& \textbf{68.41} $\pm$ 0.85
& \textbf{74.60} $\pm$ 0.99 \\
Sampling rate
& 8 fps
& 66.37 $\pm$ 0.40
& 72.26 $\pm$ 0.58 \\
\bottomrule
\end{tabular}
}
\end{table}

\paragraph{Answer self-consistency.}
Table~\ref{tab:asc_ablation} compares single-pass inference with ASC under different decoding temperatures. 
ASC improves over single-pass inference by aggregating final answer choices across multiple runs. 
We also find that stochastic decoding with temperature $1.0$ outperforms temperature $0.0$ on both validation and test, suggesting that useful answer diversity helps recover correct predictions.

\begin{table}[t]
\centering
\caption{
Effect of answer self-consistency and decoding temperature. 
Single-pass scores are averaged over 5 runs.
ASC aggregates final answer choices over $K=10$ watch-and-answer runs.
}
\label{tab:asc_ablation}
\resizebox{\linewidth}{!}{
\begin{tabular}{lcccc}
\toprule
Method
& Val AvgAcc
& Val MacroAvgAcc
& Test AvgAcc
& Test MacroAvgAcc \\
\midrule
Single pass, $T=0.0$
& 68.41
& 74.60
& 72.71
& 73.62 \\
ASC, $K=10$, $T=0.0$
& 69.13
& 74.44
& 76.80
& 75.07 \\
ASC, $K=10$, $T=1.0$
& \textbf{72.73}
& \textbf{78.34}
& \textbf{81.16}
& \textbf{80.91} \\
\bottomrule
\end{tabular}
}
\end{table}

\subsection{Additional Results}

We provide additional results in Appendix~\ref{app:category_results}, including full category-wise validation scores across methods and models. 
The complete prompts used in our pipeline are provided in Appendix~\ref{app:prompts}. 
In the next section, we analyze the main trends observed in the experiments, focusing on vote margins, recoverable uncertainty, category-wise behavior, and cross-model behavior.
\section{Analysis}
\label{sec:analysis}

We further analyze why answer self-consistency improves performance and why the effect of re-arbitration does not transfer monotonically from validation to test. 
We focus on vote margin, recoverability, low-margin category distribution, and category-wise behavior.

\subsection{Vote Margin and Recoverable Uncertainty}

We first analyze the vote margin after first-stage ASC. 
Let $m=c_1-c_2$ denote the margin between the largest and second-largest vote counts. 
A small margin means that multiple candidate answers are repeatedly selected across stochastic runs, indicating uncertainty under the current inference configuration.

Table~\ref{tab:margin_recoverability} shows the accuracy and recoverability of cumulative margin subsets on the validation split. 
Low-margin examples are much harder than high-margin examples: the subset with $m \leq 1$ has an ASC accuracy of 39.4\%, whereas the high-margin group with $m \geq 5$ reaches 81.0\%. 
This confirms that vote margin is a useful signal for identifying contested cases.

At the same time, the ground-truth answer often remains in the retained candidate set. 
For $m \leq 1$, GT Retention is 84.8\%, and it remains above 84\% for all tested low-margin thresholds. 
Thus, many low-margin errors are not cases where the correct answer is absent from the sampled answer space; rather, the correct answer is often among the top candidates but is not selected reliably by majority voting. 
This motivates MQRA, which focuses additional inference on distinguishing the retained candidates.

High GT Retention also shows that candidate inclusion alone is not sufficient. 
As the threshold expands, more ambiguous examples are triggered, and not all of them can necessarily be resolved by re-arbitration. 
Thus, a larger threshold does not necessarily lead to better performance.

\begin{table}[t]
\centering
\caption{
Recoverability of low-margin examples on the validation split after first-stage ASC.
The subset $m \leq M$ corresponds to examples triggered by MQRA with threshold $M$.
ASC Acc. is the accuracy of the ASC majority answer within each subset.
GT Retention denotes the fraction of examples where the ground-truth answer is included in the retained candidate set.
All scores are reported as percentages.
}
\label{tab:margin_recoverability}
\resizebox{\linewidth}{!}{
\begin{tabular}{lccc}
\toprule
Subset
& \# Examples
& ASC Acc.
& GT Retention \\
\midrule
$m \leq 0$
& 36
& 41.7
& 88.9 \\
$m \leq 1$
& 66
& 39.4
& 84.8 \\
$m \leq 2$
& 145
& 49.7
& 87.6 \\
$m \leq 3$
& 171
& 49.1
& 86.5 \\
$m \leq 4$
& 252
& 48.0
& 84.1 \\
\midrule
$m \geq 5$
& 749
& 81.0
& 89.6 \\
All
& 1001
& 72.7
& 88.2 \\
\bottomrule
\end{tabular}
}
\end{table}

\subsection{Split and Low-Margin Distribution}

Although low-margin examples are often recoverable, MQRA is applied only to the subset triggered by the margin threshold. 
Its effectiveness therefore depends on which types of examples appear in the triggered subset.

\begin{figure}[t]
    \centering
    \includegraphics[width=\linewidth]{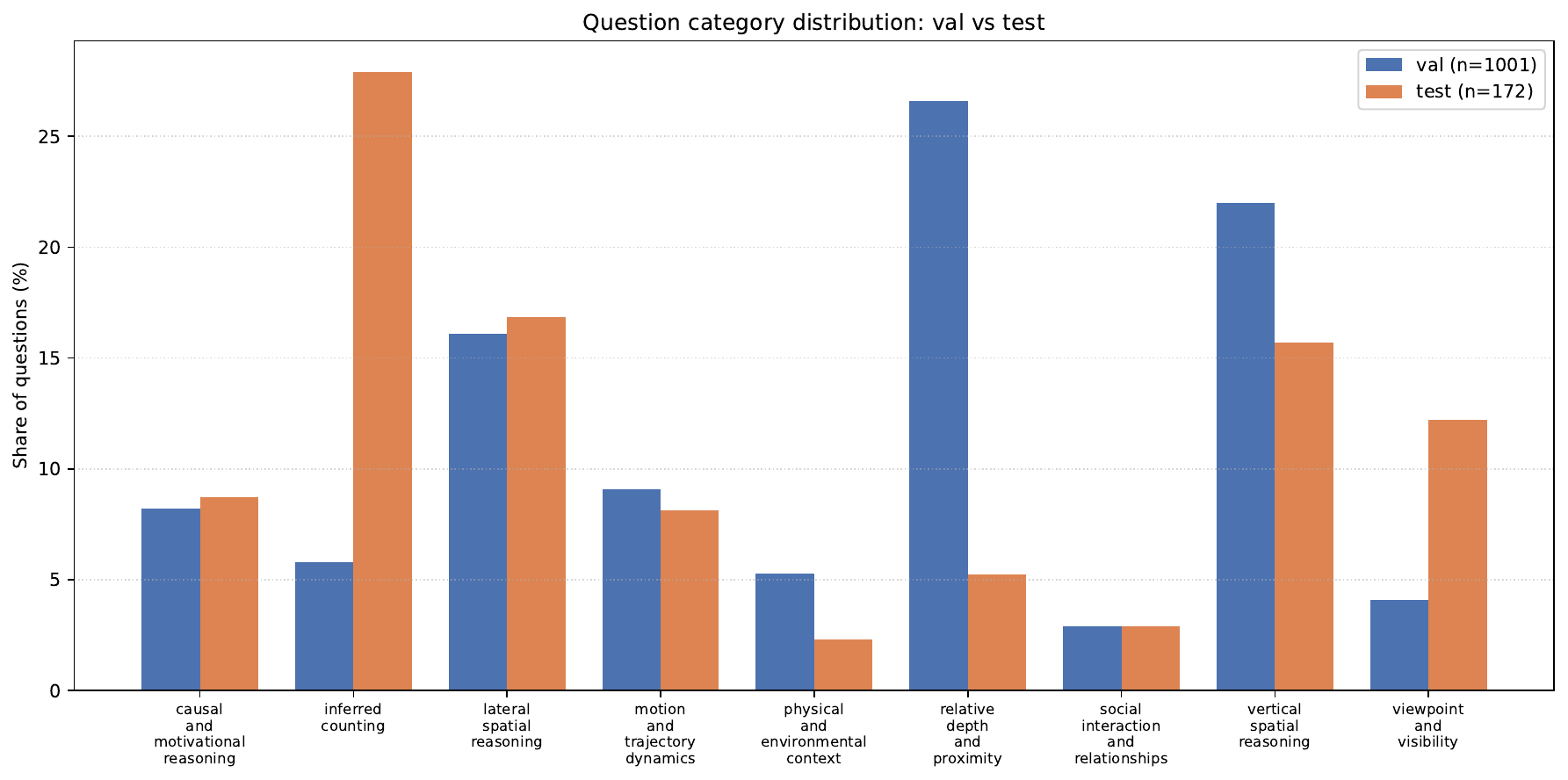}
    \caption{
    Category distribution of validation and test examples. 
    Validation uses official category labels, while test categories are estimated from the question text and answer choices for diagnostic purposes only.
    }
    \label{fig:category_dist_val_vs_test}
\end{figure}

Figure~\ref{fig:category_dist_val_vs_test} compares the category distribution of the validation and test splits. 
For validation, we use the official category labels. 
Since official category labels are not released for test, we estimate test categories from the question text and answer choices using Gemini. 
These labels are used only for diagnostic analysis and not for prediction or official evaluation.

The validation and test splits do not have identical category compositions, and low-margin examples are not uniformly distributed across categories. 
Table~\ref{tab:low_margin_category_dist} reports three complementary statistics: the number of triggered examples, the low-margin rate within each category, and the category share within the low-margin subset. 
On validation, low-margin examples are numerically concentrated in relative depth and proximity and vertical spatial reasoning, while inferred counting has the highest within-category low-margin rate. 
In contrast, the diagnostic test categories show that the triggered examples are concentrated in inferred counting and viewpoint and visibility.

\begin{table}[t]
\centering
\caption{
Category-wise distribution of low-margin examples after first-stage ASC.
``Count'' is the number of examples satisfying $m \leq 1$.
``Rate'' is the percentage within all examples of that category.
``Share'' is the percentage within the low-margin subset of that split.
Validation uses official category labels, while test categories are estimated from the question text and answer choices for diagnostic purposes only.
Categories are sorted by the number of low-margin validation examples.
}
\label{tab:low_margin_category_dist}
\resizebox{\linewidth}{!}{
\begin{tabular}{lccc|ccc}
\toprule
Category
& \multicolumn{3}{c|}{Validation}
& \multicolumn{3}{c}{Test} \\
\cmidrule(lr){2-4}
\cmidrule(lr){5-7}
& Count
& Rate
& Share
& Count
& Rate
& Share \\
\midrule
Relative depth and proximity
& 25
& 9.4\%
& 37.9\%
& 0
& 0.0\%
& 0.0\% \\
Vertical spatial reasoning
& 14
& 6.4\%
& 21.2\%
& 0
& 0.0\%
& 0.0\% \\
Inferred counting
& 11
& 19.0\%
& 16.7\%
& 6
& 12.5\%
& 50.0\% \\
Lateral spatial reasoning
& 8
& 5.0\%
& 12.1\%
& 2
& 6.9\%
& 16.7\% \\
Motion and trajectory dynamics
& 5
& 5.5\%
& 7.6\%
& 0
& 0.0\%
& 0.0\% \\
Viewpoint and visibility
& 2
& 4.9\%
& 3.0\%
& 3
& 14.3\%
& 25.0\% \\
Causal and motivational reasoning
& 1
& 1.2\%
& 1.5\%
& 1
& 6.7\%
& 8.3\% \\
Physical and environmental context
& 0
& 0.0\%
& 0.0\%
& 0
& 0.0\%
& 0.0\% \\
Social interaction and relationships
& 0
& 0.0\%
& 0.0\%
& 0
& 0.0\%
& 0.0\% \\
\midrule
Total
& 66
& 6.6\%
& 100.0\%
& 12
& 7.0\%
& 100.0\% \\
\bottomrule
\end{tabular}
}
\end{table}

These observations help explain the gap between validation and test behavior. 
On validation, MQRA improves performance for low-margin settings, suggesting that the triggered examples contain recoverable uncertainty. 
On test, however, only a small number of examples are triggered, and their category composition differs from validation. 
As a result, a few successes or failures within this triggered subset can noticeably affect the difference between ASC and ASC-MQRA.

\subsection{Category-wise Behavior}

We focus on validation for category-wise performance analysis because official category labels are available. 
Table~\ref{tab:category_wise_val} reports category-wise validation performance for single-pass inference, ASC, and ASC-MQRA with $M=0$.

ASC improves most categories, indicating that answer aggregation is broadly effective when the model can sometimes reach the correct reasoning path but does not do so consistently in a single run. 
The largest gain appears in motion and trajectory dynamics, followed by viewpoint and visibility, lateral spatial reasoning, and relative depth and proximity. 
The main exception is inferred counting, where ASC decreases performance relative to single-pass inference.

MQRA has a more selective effect. 
Compared with ASC, it mainly adds improvement in relative depth and proximity and partially recovers inferred counting, while leaving most other categories unchanged. 
This is consistent with Table~\ref{tab:low_margin_category_dist}, where relative depth and proximity has the largest share of low-margin validation examples.

\begin{table}[t]
\centering
\caption{
Category-wise validation performance.
Scores are reported as percentages.
Symbols indicate changes relative to single-pass inference:
$\uparrow$ improvement, $\downarrow$ degradation, and -- no change.
ASC-MQRA uses margin threshold $M=0$.
}
\label{tab:category_wise_val}
\resizebox{\linewidth}{!}{
\begin{tabular}{lccc}
\toprule
Category
& Single pass
& ASC
& ASC-MQRA \\
\midrule
Causal and motivational reasoning
& 92.19
& 96.34 $\uparrow$
& 96.34 $\uparrow$ \\
Inferred counting
& 59.31
& 55.17 $\downarrow$
& 56.90 $\downarrow$ \\
Lateral spatial reasoning
& 55.16
& 60.87 $\uparrow$
& 60.87 $\uparrow$ \\
Motion and trajectory dynamics
& 71.43
& 82.42 $\uparrow$
& 82.42 $\uparrow$ \\
Physical and environmental context
& 87.17
& 90.57 $\uparrow$
& 90.57 $\uparrow$ \\
Relative depth and proximity
& 56.39
& 60.53 $\uparrow$
& 62.03 $\uparrow$ \\
Social interaction and relationships
& 100.00
& 100.00 --
& 100.00 -- \\
Vertical spatial reasoning
& 75.09
& 78.64 $\uparrow$
& 78.64 $\uparrow$ \\
Viewpoint and visibility
& 74.63
& 80.49 $\uparrow$
& 80.49 $\uparrow$ \\
\bottomrule
\end{tabular}
}
\end{table}

\subsection{Model-Level Behavior}

We also examine whether the proposed test-time scaling strategy behaves similarly across different Gemini variants. 
Table~\ref{tab:cross_model_summary} summarizes the cross-model results on validation AvgAcc, with full category-wise results provided in Appendix~\ref{app:category_results}. 
The largest gain is observed for Gemini 3.1 Pro Preview, followed by Gemini 2.5 Pro. 
In contrast, Gemini 3.5 Flash and Gemini 3.1 Flash-Lite show only marginal gains.

These results suggest that the proposed pipeline is capacity-amplifying rather than capacity-replacing. 
When the base model lacks sufficient video understanding or implicit reasoning capability, additional sampling may not produce correct reasoning paths to aggregate. 
This is consistent with the small gain observed for Gemini 3.1 Flash-Lite. 
At the same time, a high single-pass score alone does not guarantee a large gain from test-time scaling. 
Gemini 3.5 Flash achieves a relatively strong single-pass score, but its repeated baseline runs have much smaller standard deviation than Gemini 3.1 Pro Preview, suggesting that its predictions are already more stable and may leave less room for aggregation-based improvement.

A useful condition appears to be \emph{recoverable uncertainty}: the model has enough capability to sometimes reach the correct answer, but does not select it consistently in a single pass. 
In this setting, prediction variability may reflect not only random instability but also useful diversity among plausible reasoning paths. 
ASC-MQRA can then stabilize this diversity through answer-level aggregation and focused re-arbitration. 
We do not claim to observe the internal reasoning process of each model, but the cross-model results suggest that both base capability and useful prediction diversity affect the benefit of test-time scaling.

\begin{table}[t]
\centering
\caption{
Cross-model behavior of ASC-MQRA on the validation split.
Scores are validation average accuracy (Val AvgAcc) reported as percentages.
ASC-MQRA uses margin threshold $M=0$ for all models.
Full category-wise results are provided in Appendix~\ref{app:category_results}.
}
\label{tab:cross_model_summary}
\resizebox{\linewidth}{!}{
\begin{tabular}{lccc}
\toprule
Model
& Single pass
& ASC-MQRA
& Gain \\
\midrule
Gemini 3.1 Pro Preview
& 68.41
& 73.23
& +4.82 \\
Gemini 2.5 Pro
& 63.68
& 65.83
& +2.15 \\
Gemini 3.5 Flash
& 70.31
& 70.73
& +0.42 \\
Gemini 3.1 Flash-Lite
& 57.34
& 57.54
& +0.20 \\
\bottomrule
\end{tabular}
}
\end{table}

\subsection{Practical Observations and Limitations}

ASC often helps when different runs attend to different parts of the video and at least some runs identify the decisive cue. 
In such cases, answer aggregation can suppress isolated mistakes and recover the answer supported by the most stable reasoning pattern. 
MQRA can further help when disagreement is concentrated among a small number of plausible candidates, because the arbitration prompt explicitly asks the model to clarify the question and distinguish the remaining options.

However, both ASC and MQRA have limitations. 
If all sampled runs miss the same decisive cue, answer aggregation cannot recover the correct answer. 
MQRA can also hurt when the correct answer is not included in the retained candidate set, or when the arbitration prompt over-focuses on a misleading candidate difference. 
These failure modes are consistent with the retention analysis in Table~\ref{tab:margin_recoverability}. 
\section{Conclusion}
\label{sec:conclusion}

In this report, we presented ASC-MQRA, a training-free test-time reasoning pipeline for Track 2 of the CVPR 2026 VidLLMs Challenge. 
Our method improves video question answering by sampling multiple watch-and-answer runs and aggregating their answer choices through answer-level self-consistency. 
This ASC component substantially improves over single-pass inference on both validation and test, achieving strong performance on both AvgAcc and MacroAvgAcc. 
We further studied margin-triggered question re-arbitration, which allocates additional inference to low-margin examples. 
MQRA improves validation performance when the correct answer remains recoverable among the retained candidates, but does not improve test performance under a different triggered-category distribution.

Our analysis suggests that the proposed approach is most effective under recoverable uncertainty: cases where the model can sometimes reach the correct answer, but does not select it consistently in a single pass. 
Vote-margin, category-wise, and cross-model analyses indicate that test-time scaling amplifies the capability of a sufficiently strong base model rather than replacing missing video reasoning ability. 
At the same time, the effect of re-arbitration depends on the category composition and recoverability of the triggered examples, which helps explain the difference between validation and test behavior.

Several limitations remain. 
Difficult categories such as inferred counting, lateral spatial reasoning, and relative depth and proximity remain challenging, especially when questions require robust spatial reconstruction, viewpoint tracking, object orientation, or reasoning across abrupt scene changes. 
The pipeline also increases inference cost because it relies on multiple video question-answering runs and additional re-arbitration for low-margin examples. 
Future work should explore more reliable spatial-temporal reasoning strategies and more efficient uncertainty-aware test-time scaling. 
Overall, our results show that simple answer-level test-time scaling can be a strong training-free strategy for implicit video reasoning.

{
    \small
    \bibliographystyle{ieeenat_fullname}
    \bibliography{main}
}

\clearpage
\onecolumn
\appendix
\section*{Appendix}
\section{Prompts}
\label{app:prompts}

We provide the prompts used in the first-stage watch-and-answer runs and the margin-triggered re-arbitration stage. 
The first-stage prompt asks the model to choose a single answer and provide video-grounded evidence. 
The arbitration prompt is used only for low-margin examples and asks the model to clarify the question, distinguish the retained candidates, and re-watch the video before selecting one candidate.

\subsection{First-Stage Watch-and-Answer Prompt}

\promptheading{System prompt.}
\begin{lstlisting}[style=promptstyle]
You are a careful visual reasoning expert answering a multiple-choice question about a video.

The deciding evidence is often implicit: it may require connecting events across distant moments, reading subtle visible cues, tracking who did what, or noticing small state changes. Do not answer based on what is merely plausible or what sounds like a typical answer.

For your answer you must fill in:
- "answer_choice": your single best answer (an option letter or number). ALWAYS provide exactly one.
- "evidence": the concrete, visually-grounded observation(s) that support "answer_choice". Cite what is visible and roughly when. If your evidence is weak, say so honestly.

Answer only from what is visible in the video. Do not invent details.
\end{lstlisting}

\promptheading{User prompt.}
\begin{lstlisting}[style=promptstyle]
Question:
{question_text}

Answer options:
{answer_option_text}

Watch the video carefully and choose the single best answer.

The answer may require connecting visible events across the video. Do not choose an option only because it mentions a visually salient person, animal, object, or event. Answer only from what is visible.

Return your structured answer with "answer_choice" and "evidence".
\end{lstlisting}

\subsection{Re-Arbitration Prompt}

\promptheading{System prompt.}
\begin{lstlisting}[style=promptstyle]
You are a careful visual reasoning expert making the final decision on a multiple-choice question about a video whose answer is contested. You are given the question and the remaining candidate options. Re-watch the video and choose the candidate that the video best supports.

Decide which candidate option is correct based on what is visible (or strongly implied) in the video and how directly it answers the question -- NOT on prior votes or candidate preferences.

For your answer you must fill in "question_understanding", "candidate_difference", "answer_choice", and "evidence". "answer_choice" MUST be one of the remaining candidate options. Answer only from what is visible in the video. Do not invent details.
\end{lstlisting}

\promptheading{User prompt.}
\begin{lstlisting}[style=promptstyle]
You are resolving a disagreement between candidate answers for a multiple-choice video question.

Question:
{question_text}

Remaining candidate options (choose exactly one of these):
{candidate_option_text}

Watch the video again carefully and choose the single best answer from the remaining candidate options.

Follow these steps carefully:

1. Understand the question correctly.
   Restate what the question is asking in your own words.
   Be precise about what must be answered: a person/object, count, order, direction, spatial relation, viewpoint, cause/reason, outcome/result, identity, social relation, or physical state.
   Mention only the aspects that are relevant to this question.
   Do not add assumptions that are not required by the question.

2. Clarify the difference between the remaining candidates.
   Identify what makes the candidate options different from each other.
   Focus on the decisive visual fact, relation, event, or inference that would make one candidate correct and the others wrong.
   Do not merely restate the original question.

3. Re-watch the video and judge.
   Verify what is actually visible or strongly implied by the video.
   Treat unsupported assumptions as unreliable.
   Do not use earlier vote counts or prior candidate preferences.

4. Answer the question.
   Choose exactly one remaining candidate.
   Select the option that directly answers the question and is best supported by the video-grounded evidence.
   If evidence is still incomplete, choose the option that requires the fewest unsupported assumptions.

Return JSON only:
{{
  "question_understanding": "...",
  "candidate_difference": "...",
  "answer_choice": "...",
  "evidence": "..."
}}

Rules:
- Choose exactly one option from the remaining candidate options.
- Do not introduce a new option.
- Do not decide by vote count.
- Do not answer a different question than the one asked.
- If the question asks why, answer the cause/reason.
- If the question asks what happened, answer the outcome/result.
- If the question specifies a viewpoint or reference, use that viewpoint/reference.
- If the question asks for a count or order, track the relevant entities/events across the required scope.
- Base the answer only on the video and grounded evidence.
\end{lstlisting}

\subsection{Question Category Classification Prompt}

Official category labels are not released for the test split. 
For diagnostic analysis only, we estimated test categories from the question text and answer choices using the following prompt. 
The estimated categories were not used for prediction or official evaluation.

\promptheading{User prompt.}
\begin{lstlisting}[style=promptstyle]
You are classifying a multiple-choice question about a video into one VRR (Video Reasoning) category.
You will NOT see the video. Use only the question and the answer options.

Question:
{question_text}

Answer options:
{answer_option_text}

Choose exactly one category from the following list:

- lateral_spatial_reasoning:
  Inferring left/right or side-to-side spatial relationships between objects or characters, often across scenes, cuts, viewpoints, or incomplete views.

- vertical_spatial_reasoning:
  Inferring above/below or vertical arrangements between objects or characters.

- relative_depth_and_proximity:
  Inferring which objects or characters are closer, farther, nearer to each other, or separated in depth.

- viewpoint_and_visibility:
  Inferring what can be seen from a camera angle or character viewpoint, including line-of-sight, occlusion, and visibility.

- motion_and_trajectory_dynamics:
  Inferring movement direction, path, or trajectory within or across scenes, especially when the full relation is not directly shown in one scene.

- causal_and_motivational_reasoning:
  Inferring causes, intentions, motivations, or likely consequences from incomplete visual cues.

- inferred_counting:
  Counting or ordering entities/events by integrating scattered evidence across scenes, rather than counting from a single visible moment.

- physical_and_environmental_context:
  Inferring physical continuity, object state, environmental layout, or scene context that is implied but not directly shown.

- social_interaction_and_relationships:
  Inferring relationships, roles, or social dynamics between characters from behavior, reactions, or contextual cues.

Return JSON only in this format:
{{"category": "<one of the listed VRR categories>"}}
Do not explain.
\end{lstlisting}

\clearpage
\section{Additional Category-wise Results}
\label{app:category_results}

Table~\ref{tab:category_results_full} reports category-wise results on the official validation split across methods and models. 
Scores are reported as percentages, and ASC-MQRA uses margin threshold $M=0$ for all models.

\begin{table}[t]
\centering
\caption{
Category-wise validation results across methods and models.
Scores are reported on the official validation split as percentages.
ASC-MQRA uses margin threshold $M=0$ for all models.
}
\label{tab:category_results_full}
\scriptsize
\resizebox{\linewidth}{!}{
\begin{tabular}{llccccccccccc}
\toprule
Model
& Method
& \shortstack{Causal and\\motivational\\reasoning}
& \shortstack{Inferred\\counting}
& \shortstack{Lateral\\spatial\\reasoning}
& \shortstack{Motion and\\trajectory\\dynamics}
& \shortstack{Physical and\\environmental\\context}
& \shortstack{Relative depth\\and proximity}
& \shortstack{Social interaction\\and relationships}
& \shortstack{Vertical\\spatial\\reasoning}
& \shortstack{Viewpoint\\and visibility}
& AvgAcc
& \shortstack{Macro\\AvgAcc} \\
\midrule
\multirow{2}{*}{Gemini 3.1 Pro Preview}
& Single pass
& 92.19
& 59.31
& 55.16
& 71.43
& 87.17
& 56.39
& 100.00
& 75.09
& 74.63
& 68.41
& 74.60 \\
& ASC-MQRA
& 96.34
& 56.90
& 60.87
& 82.42
& 90.57
& 62.03
& 100.00
& 78.64
& 80.49
& 73.23
& 78.69 \\
\midrule
\multirow{2}{*}{Gemini 2.5 Pro}
& Single pass
& 90.00
& 46.55
& 51.31
& 72.53
& 87.93
& 50.08
& 92.41
& 70.45
& 64.39
& 63.68
& 69.52 \\
& ASC-MQRA
& 92.68
& 50.00
& 53.42
& 68.13
& 88.68
& 53.38
& 93.10
& 73.64
& 68.29
& 65.83
& 71.26 \\
\midrule
\multirow{2}{*}{Gemini 3.5 Flash}
& Single pass
& 91.46
& 54.48
& 59.01
& 81.76
& 88.68
& 61.35
& 89.66
& 73.91
& 70.73
& 70.31
& 74.56 \\
& ASC-MQRA
& 92.68
& 62.07
& 54.66
& 84.62
& 86.79
& 62.41
& 96.55
& 73.64
& 70.73
& 70.73
& 76.02 \\
\midrule
\multirow{2}{*}{Gemini 3.1 Flash-Lite}
& Single pass
& 82.69
& 33.45
& 44.22
& 65.93
& 82.27
& 46.32
& 86.21
& 62.45
& 64.39
& 57.34
& 63.10 \\
& ASC-MQRA
& 84.15
& 36.21
& 42.86
& 65.93
& 83.02
& 46.99
& 89.66
& 62.73
& 58.54
& 57.54
& 63.34 \\
\bottomrule
\end{tabular}
}
\end{table}

\clearpage
\section{Alternative Directions Considered}
\label{app:unsuccessful_directions}

We also experimented with or considered several alternative prompting and aggregation strategies that were not included in the final pipeline. 
Some were empirically tested but did not consistently improve validation performance, while others were considered based on related work but appeared less suitable for our setting. 
We include them here to clarify why the final pipeline uses answer-level aggregation with margin-triggered re-arbitration.

\subsection{Evidence-Level Aggregation}

We experimented with aggregating generated evidence or observations across multiple runs before producing the final answer. 
The motivation was to use agreement and contradiction among intermediate explanations as an additional signal for resolving uncertain predictions. 
However, generated evidence was often noisy or inconsistent on difficult VRR-QA examples. 
In some cases, incorrect observations were repeated across runs or became over-emphasized during aggregation. 
As a result, evidence-level aggregation did not provide stable improvements.

This motivated our final design choice: ASC aggregates only final answer choices, while the evidence field is used only to encourage grounded reasoning within each individual run and to support qualitative inspection.

\subsection{Spatial Cognitive-Map Prompting}

We experimented with prompts that encourage the model to construct an internal spatial map before answering spatial reasoning questions. 
This direction was motivated by recent work on spatial reasoning and cognitive maps in multimodal models~\cite{yang2024think}. 
The goal was to make the model explicitly reason about relative positions, visibility, depth, movement, and viewpoint before selecting an answer.

However, this strategy did not consistently improve validation performance. 
VRR-QA often includes abrupt shot changes, camera motion, viewpoint shifts, vertical relations, depth relations, and orientation-dependent questions involving people, characters, or objects. 
These factors make it difficult to represent the relevant scene structure with a simple prompt-induced spatial map. 
We found that prompt-level spatial reasoning alone was not sufficient to reliably improve categories such as lateral spatial reasoning, relative depth and proximity, and vertical spatial reasoning.

\subsection{Category-Specific Rule Prompting}

We also experimented with category-specific prompts that added explicit reasoning rules for particular question types. 
For example, for spatial reasoning questions, we prompted the model to identify the target object, the reference object, and the requested relation before selecting an answer. 
The prompt also included rules for interpreting left/right, above/below, same/opposite side, and viewpoint-dependent relations.

This type of prompting sometimes improved individual examples by making the model attend to the relevant relation more explicitly. 
However, the effect was not stable across the validation set. 
In some cases, the additional rules helped resolve ambiguity; in others, they caused the model to over-structure the question, focus on an incorrect target or reference, or apply a rule that did not match the visual evidence. 
As a result, category-specific rule prompting did not provide a reliable overall improvement and was not included in the final pipeline.

\subsection{Confidence-Aware Efficient Self-Consistency}

Because ASC requires multiple video question-answering runs, we considered more efficient confidence-aware self-consistency methods. 
Recent work suggests that confidence estimates can improve self-consistency or reduce the number of sampled reasoning paths~\cite{taubenfeld2025confidence}. 
Such methods are attractive for our setting because video inference is expensive.

However, the most relevant variants require reliable confidence or probability signals for each sampled answer. 
In our implementation setting, integrating such signals with Gemini-based video inference was not straightforward. 
In particular, token-level probability access and video-input handling introduced additional practical constraints. 
We therefore used a simpler answer-level self-consistency approach with a fixed number of runs as a robust alternative.

\subsection{Moment-Centric Sampling}

We also considered moment-centric or localization-style strategies that first identify a small set of relevant temporal segments before answering. 
Moment selection is attractive for long-form video question answering, and recent work has shown that moment retrieval can guide frame sampling for long-form video question answering~\cite{chasmai2025moment}. 
This direction could reduce the amount of video context processed by the model and focus inference on likely decisive evidence.

In VRR-QA, however, many questions are not answered by a single decisive moment. 
They often require connecting multiple separated cues across the video, such as tracking an entity across shots, inferring a relation from before-and-after states, or combining spatial and contextual information. 
For this reason, we focused on full-video watch-and-answer runs rather than a moment-only pipeline.

\end{document}